\documentclass{article}
\usepackage{spconf,amsmath,epsfig}
\usepackage{amsfonts}
\usepackage{amssymb}
\usepackage{lineno}
\usepackage{tabularx}
\usepackage{multirow}
\usepackage{multicol}
\usepackage{subfig}
\usepackage{array}
\usepackage{graphicx}
\usepackage{caption}
\usepackage{tikz}
\usepackage{xspace}
\usepackage{enumerate}
\usepackage{xurl}
\usepackage[colorlinks,allcolors=blue]{hyperref}

\usepackage{colortbl}
\newcommand{\myrowcolour}{\rowcolor[gray]{0.94}}
\newcommand{\highest}[1]{{\underline{${#1}$}}}%

\usepackage[normalem]{ulem} 

\title{A Comparative Assessment of Multi-view fusion learning for Crop Classification}
\name{{Francisco Mena$^{1,2}$, Diego Arenas$^{2}$, Marlon Nuske$^{2}$, and Andreas Dengel$^{1,2}$}
\thanks{F. Mena acknowledges the financial support from the chair of Prof. Dr. Prof. h.c. Andreas Dengel with RPTU in Kaiserslautern.}}
\address{
\begin{minipage}[c]{0.44\textwidth}
\centering
$^{1}$University of Kaiserslautern-Landau (RPTU)\\
Department of Computer Science\\
Kaiserslautern, Germany\\
\end{minipage}
\quad
\begin{minipage}[c]{0.57\textwidth}
\centering
$^{2}$German Research Center for Artificial Intelligence (DFKI)\\
Smart Data and Knowledge Services\\
Kaiserslautern, Germany
\end{minipage}
}

\begin{document}
%
\maketitle
\begin{abstract} 
With a rapidly increasing amount and diversity of remote sensing (RS) data sources, there is a strong need for multi-view learning modeling.
This is a complex task when considering the differences in resolution, magnitude, and noise of RS data. 
The typical approach for merging multiple RS sources has been input-level fusion, but other - more advanced - fusion strategies may outperform this traditional approach. 
This work assesses different fusion strategies for crop classification in the CropHarvest dataset. The fusion methods proposed in this work outperform models based on individual views and previous fusion methods. 
We do not find one single fusion method that consistently outperforms all other approaches. Instead, we present a comparison of multi-view fusion methods for three different datasets and show that, depending on the test region, different methods obtain the best performance. Despite this, we suggest a preliminary criterion for the selection of fusion methods.
\end{abstract}

\section{Introduction} \label{sec:intro}

Many phenomena in our environment are studied through multiple sources, e.g. a farm that could be observed by satellites with different sensors. The idea is to corroborate and complement the information between observations. 
Deep learning models have been widely used to model complex relationships between input data and target tasks. However, the situation described above poses a machine learning scenario of combining information coming from multiple sources, named multi-view or multi-modal fusion learning. 
In the context of Remote Sensing (RS), relevant challenges arise regarding the different types of sensor resolution.
For instance, RS views could have different spatial or temporal resolutions, even different spectral bands (bandwidth or numbers of channels), calibration, and amount of noise.

In RS, there could be cases when a model trained with individual views (single-view model) gives predictions similar to a random classifier. See AA scores in Tab.~\ref{tab:single_random} of DEM and Weather views for an example. We remark this as the main difference between multi-view learning in RS \cite{mena2023} to conventional multi-modal learning \cite{yan2021deepmultiviewlearning} (e.g. with image, text, and audio).
This suggests that individual RS views may not contain enough information to achieve an optimal classification for some tasks, and that multi-view fusion is required.
\begin{table}[!t]
    \centering
    \small
    \begin{tabular}{cc|ccc}\hline
        Data & View & AA & AUC & Entropy \\
        \hline
        \myrowcolour \multirow{3}{*}{Kenya} & DEM & $48.0 \pm 0.9$ & $41.1 \pm 0.9$ & $83.1 \pm3.8$ \\
                             & Weather  & $50.0 \pm 0.0$ & $37.1 \pm 0.0$ & $72.2 \pm8.7$  \\
        \myrowcolour             & \textbf{Radar} & ${63.0} \pm 1.0$ & ${66.8} \pm 2.9$ & $69.2 \pm3.8$ \\ \hline
         & DEM & $61.2 \pm 2.6$ & $65.0 \pm 1.8$ & $94.6\pm1.3$  \\
        \myrowcolour Togo    & Weather & $55.6 \pm 3.5$ & $59.3 \pm 5.4$ & $98.6\pm0.7$  \\ 
            & \textbf{Optical} & ${80.0} \pm 1.3$ & ${88.6} \pm 0.7$ & $62.7\pm2.1$  \\
            \hline
        \myrowcolour \multirow{3}{*}{Global} & DEM & $ 64.9\pm 0.1$ & $ 69.7\pm 0.1$ & $91.1 \pm 1.1$  \\
            & Weather & $ 72.7 \pm 0.8$ & $ 81.2\pm 0.6$ & $ 74.1\pm 1.4$  \\ 
        \myrowcolour & \textbf{Optical} & $ 78.7 \pm 0.8$ & $ 86.7 \pm 0.8$ & $ 59.6 \pm 2.0$  \\
            \hline
    \end{tabular}
    \caption{Predictive performance of a model trained with individual views. The view with the best individual performance for the task is highlighted in \textbf{bold}.}
    \label{tab:single_random}
\end{table} 

The most common approach used in the literature to merge the information on multi-view RS is the input-level fusion \cite{ofori-ampofo2021croptypemappingb,tseng2021cropharvestglobaldataset}, i.e. align the input views to the same resolution (e.g. spatial and temporal interpolation) and concatenate them. 
However, some works have shown that the heterogeneity of RS views could be better exploited by using other fusion approaches \cite{ofori-ampofo2021croptypemappingb,hong2021morediversemeansa}. 
In this work, we compare different multi-view fusion methods for the crop classification task. 
The key contributions are as follows:
\begin{itemize}
    \setlength\itemsep{0em}
    \item A new assessment of different fusion strategies on the recent CropHarvest dataset \cite{tseng2021cropharvestglobaldataset}.
    \item Achieve state-of-the-art results without pre-training in the African testing regions (Kenya and Togo).
    \item Open-source code to ensure reproducibility of our experiments and findings.\footnote{https://github.com/fmenat/MultiviewCropClassification}.
\end{itemize}

\section{Case Study} \label{sec:case}

For this case study, we use a recent benchmark dataset on crop classification called CropHarvest \cite{tseng2021cropharvestglobaldataset}. 
Each sample corresponds to a yearly time-series at the pixel-level. The samples are labeled as positive or negative for specific crop categories, i.e. the task corresponds to identifying whether a pixel contains a specific (or any) growing crop during a season.
There are 5 views as input data coming from 4 RS sources: Sentinel-2, Sentinel-1, ERA5, and SRTM.
The multi-band optical view with 11 spectral channels was obtained from Sentinel-2 at 10-60m/px spatial resolution and 5 days temporal resolution. 
The 2-band radar view (VV and VH) was obtained from Sentinel-1 at 10m/px and a variable revisit time.
The 2-band weather view (precipitation and temperature) was obtained from ERA5 at 31km/px and hourly temporal resolution.
The 1-band Normalized Difference Vegetation Index (NDVI) was calculated from the optical view.
The 2-band Digital Elevation Map (DEM) view (elevation and slope) was obtained from SRTM at 30m/px and contains topography information (static across time).
The views are aligned by monthly averaging (temporal) and 10-m interpolating (spatial).

Out of the total 95186 samples in the benchmark, we only use the 65243 samples that have all views available. The dataset contains two geographical testing regions in Africa: Kenya and Togo.
We also include a global testing region, which consists of all the training samples across the globe.
Tab.~\ref{tab:data} displays the distribution of the samples per region.
\begin{table}[!ht]
    \centering
    \begin{tabular}{c|ccc} \hline
        Data & Total & Training & Testing \\ \hline
        \myrowcolour Kenya &  2217 ($37.8\%$) & 1319 ($20.0\%$) & 898 ($64.0\%$) \\
        Togo  & 1596 ($51.1\%$) & 1290 ($55.0\%$) & 306 ($34.6\%$) \\
        \myrowcolour Global & 65243 ($66.3\%$) & 45723 ($66.4\%$) & 19520 ($66.0$) \\ \hline
    \end{tabular}
    \caption{Number of samples in each data region. The percentage of positive data is shown in parentheses.}
    \label{tab:data}
\end{table}

\section{Methods} \label{sec:methods}
\begin{figure*}[!t]
\centering
\begin{minipage}[b]{0.5\linewidth}
  \centering
  \includegraphics[width=\textwidth, page=1]{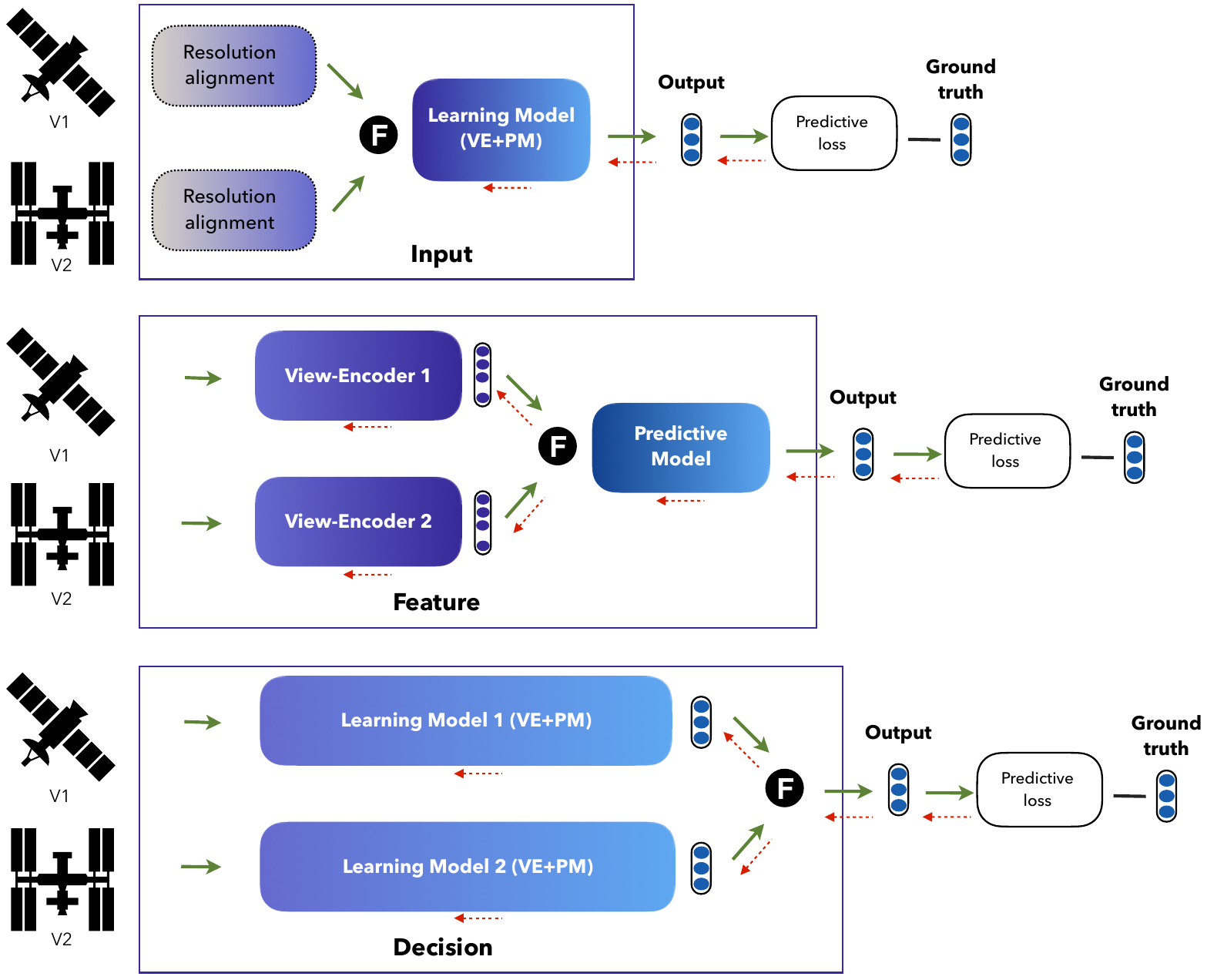}
  \centerline{(a) Three main fusion strategies.}\medskip
\end{minipage}
\quad
\begin{minipage}[b]{.47\linewidth}
  \centering
  \includegraphics[width=\textwidth, page=2]{figs/fusions_crop_class.pdf}
  \centerline{(b) Additional fusion strategies.}\medskip
\end{minipage}
\caption{Fusion methods compared in this manuscript. Green arrows represent the forward pass of the models (left to right), while red arrows represent the backward pass. VE stands for view-encoder and PM for predictive model. Images from \cite{mena2023}.} \label{fig:methods}
\end{figure*}

We compare three main fusion strategies highlighted in \cite{mena2023,ofori-ampofo2021croptypemappingb,hong2021morediversemeansa}: input, feature, and decision level fusion. In addition, multiple losses model \cite{benedetti2018textfusiondeeplearninga} and an ensemble-based aggregation \cite{robinson2021globallandcovermapping} are compared. An illustration of these methods is in Fig.~\ref{fig:methods}.

\textbf{Input}: a concatenation of \textit{input views} with a resolution alignment is directly fed to a single model.
\textbf{Feature}: it uses view-encoder models to map each view to a new \textit{feature space} (named view-representation), followed by a merge function and fully connected layers. We test different merge functions. Simple summary functions 
(\textbf{Feature-S}) and with gated modules (\textbf{Feature-G}). 
The gated module adaptively fuses the view-representations through a weighted sum, where the (attention) weights are computed for each sample \cite{arevalo2020gatedmultimodalnetworksa}. The weights could be modeled for each view or for each feature and view combination (feature-specific weight).
\textbf{Decision}: it uses parallel models that process each view, outputs the crop probability (\textit{decision}), and then averages to yield the aggregated prediction. 
\textbf{Multi-Loss}: based on the feature-level fusion, it includes one auxiliary predictive model that is fed with a single view representation and yields the crop classification. For training the auxiliary model for each view, a loss function is added to the optimization (with a weight of $0.3$, \cite{benedetti2018textfusiondeeplearninga}), so the model has to learn the task based on the individual views in addition to the fused information.
\textbf{Ensemble}: similar to decision-level fusion, but on a two-step basis. The first step corresponds to training a model for each view (without fusion), while the second step aggregates (averages) the predicted probabilities from individual models at test time.

We use a recurrent neural network with 2 GRU layers and 64 units as view-encoders. While for prediction, we stack 1 fully connected layer with 64 hidden units.
We include 20\% of dropout during training and batch-normalization on all layers, as suggested \cite{chen2017deepfusionremotea}. 
The optimization is carried out with the cross entropy loss and ADAM optimizer with a 256 batch size.
We train the models for 1000 epochs maximum with an early stopping criterion, patience of 5, and delta loss of $0.01$ on the validation set (a 10\% randomly selected from training).

\section{Experiments} \label{sec:experiments}

We use two classification metrics \cite{tseng2021cropharvestglobaldataset}, Average Accuracy (AA), and Area Under the Curve (AUC). 
We also include the entropy of the predicted crop probability to analyze the uncertainty of the model. 
Based on the variability of the weights initialization and stochastic optimization, we repeat the experiment 10 times and report the average and std. deviation.

By selecting the best combination of the merge function and the gate type for Feature methods (see Section~\ref{sec:experiments:ablation} for the corresponding experiment), we compare all fusion methods in Tab.~\ref{tab:main_comparison}.
We observe that methods using multiple models often obtain better classification performance than a single model (Input fusion).
Besides, fusions at the feature-level are always in the top 3 best performances and have the lowest std. deviation.
The evidence shows the effectiveness of multi-view fusion, since the top 3 fusion methods have better classification performance than individual views (from Tab.~\ref{tab:single_random}) in all cases. 
In terms of \textit{relative improvement}, $100(a-b)/b$, when we compare the best fusion method ($a$) with the best model on an individual view ($b$), a \textit{relative improvement} of $6\%$, $5\%$, and $4\%$ is obtained in AA for Kenya, Togo, and Global, respectively. 
While comparing the best fusion ($a$) with the worst individual model ($b$), we get a relative improvement of $39\%$, $37\%$, and $27\%$ in AA for Kenya, Togo and Global. 
These results suggest that significant improvements can be obtained by exploring different fusion strategies.

Regarding entropy, the Feature-G method obtains the lowest uncertainty in its classification predictions. It is possible that, by letting the model choose how it will combine the views for each sample (i.e. the adaptive fusion in the gated module proposal \cite{arevalo2020gatedmultimodalnetworksa}), it becomes more confident in its decision.
In most methods, fusing multiple views decreases the uncertainty regarding individual views. However, with some methods the entropy increases, e.g. with Ensemble.
Interestingly, when merging towards the output-level, the model increases the entropy, e.g. the order of increasing entropy goes like Input, Feature-S, and then Decision. This suggests that the method becomes more uncertain when it has few layers to exchange information between the views.

\begin{table}[!t]
    \centering
    \small
    \begin{tabular}{cc|ccc} \hline
        Data & Method & AA & AUC & Entropy  \\
        \hline
        \myrowcolour  & Input & $ 61.3\pm 7.5$ & $ 70.0\pm 7.2$ & $ 72.9\pm 6.1$  \\
                             & Feature-S  & \highest{63.0 \pm 5.1} & \highest{71.6 \pm 3.7} & $ 73.9 \pm 4.4$ \\
        \myrowcolour  Kenya                   & Feature-G  &  \highest{66.5 \pm 3.6} &  \highest{71.8\pm 3.5} & $ 71.6\pm 5.8$  \\
                            & Decision & $ 57.5\pm 7.6$ & $ 63.2\pm 7.0$ & $ 76.8\pm 7.6$ \\
        \myrowcolour    & Ensemble & $ 56.0 \pm 5.3$ & $ 69.5\pm 1.4$ & $ 83.5\pm 2.2$\\
                            & Multi-Loss &  \highest{64.8\pm 5.2} &  \highest{71.6\pm 4.1} & $ 73.4\pm 4.3$ \\
                     \hline
        \myrowcolour & Input & $ 79.7\pm 1.5$ &  \highest{89.0\pm 1.3} & $ 54.3\pm 4.2$  \\
                             & Feature-S & \highest{79.9 \pm 1.0} & $ 88.9 \pm 0.8$ & $ 57.6\pm 4.1$ \\
        \myrowcolour  Togo     & Feature-G  & $ 78.1\pm 1.8$ & $ 87.6\pm 1.2$ & $ 52.1\pm 7.8$  \\
                            & Decision &  \highest{81.5\pm 1.6} &  \highest{89.5\pm 0.5} & $ 63.7\pm 2.5$ \\
        \myrowcolour    & Ensemble &  \highest{84.0\pm 1.0} &  \highest{90.9\pm 0.5} & $ 93.5\pm 0.6$ \\
        & Multi-Loss & $ 78.2\pm 6.1$ & $ 88.6\pm 1.7$ & $ 60.4\pm 12.2$ \\
                     \hline
        \myrowcolour  & Input & $ 81.2 \pm 0.9$ &  \highest{89.7\pm 0.8} & $ 52.7\pm 2.5$  \\
                             & Feature-S  & \highest{81.3\pm 1.1} & $ {89.6\pm 1.1}$ & $ 53.8\pm 3.8$ \\
        \myrowcolour Global        & Feature-G  & \highest{82.1\pm 1.4} &  \highest{90.6\pm 1.1} & $ 51.1\pm 4.3$  \\
                            & Decision & $ 79.1 \pm 1.4$ & $ 87.8 \pm 1.3$ & $59.4\pm 4.6$ \\
        \myrowcolour       & Ensemble & $ 79.2\pm 0.8$ & $ 87.5\pm 1.5$ & $ 85.4\pm 0.6$ \\
                & Multi-Loss &  \highest{81.6 \pm 1.4} &  \highest{90.1\pm 1.1} & $ 52.3\pm 3.7$ \\
            \hline
    \end{tabular}
    \caption{Prediction performance of multi-view fusion methods. The top 3 best results for each test are \underline{underlined}. The mean and std. deviation between multiple runs is shown.}
    \label{tab:main_comparison}
\end{table}

\subsection{State-of-the-Art comparison} \label{sec:experiments:sota}

The Tab.~\ref{tab:sota} compares our results with the state-of-the-art including the binary F1 score. We show that, by selecting an appropriate model for fusion, we outperform previous work.
Indeed, the best method is different for each region and metric. In Kenya, we obtain the best AUC with Feature-G, and the best binary F1 with Feature-S, while in Togo with the Ensemble method.
\begin{table}[!t]
    \centering
    \small
    \begin{tabular}{cc|cc}\hline
        Data & Method &  AUC & binary F1  \\
        \hline
        \myrowcolour \multirow{3}{*}{Kenya} & Random Forest \cite{tseng2021cropharvestglobaldataset} & $57.8 \pm 0.6$ & $55.9 \pm 0.3$  \\
                             & LSTM \cite{tseng2021cropharvestglobaldataset}  & $32.9 \pm 1.1$ & $78.2 \pm 0.0$  \\
        \myrowcolour             & Feature-S & $71.6 \pm 3.7$ & $\mathbf{79.4 \pm 3.3}$ \\
                     & Feature-G & $\mathbf{71.8} \pm {3.5}$ & $77.2 \pm 4.1$ \\
                     \hline
        \myrowcolour \multirow{3}{*}{Togo} & Random Forest \cite{tseng2021cropharvestglobaldataset} & $89.2 \pm 0.1$ & $75.6 \pm 0.2$  \\
                             & LSTM \cite{tseng2021cropharvestglobaldataset}  & $86.1 \pm 0.2$ & $72.0 \pm 0.5$  \\
        \myrowcolour                    & Ensemble & $\mathbf{90.9 \pm 0.5}$ & $\mathbf{77.8 \pm 1.3}$ \\
            \hline
    \end{tabular}
    \caption{State-of-the-art prediction performance on the CropHarvest dataset. The best average result is in \textbf{bold}.}
    \label{tab:sota}
\end{table}

\subsection{Ablation} \label{sec:experiments:ablation}
Firstly, we compare different merge functions for the feature-level fusion with simple functions (Feature-S) in Tab.~\ref{tab:merge_func}. 
The \textit{average} operation has the best classification performance for Kenya and Togo, while the \textit{concatenation} has it for the Global region.
These best performing approaches also have the lowest entropy.
\begin{table}[!t]
    \centering
    \small
    \begin{tabular}{cc|ccc} \hline
        Data & Merge & AA & AUC & Entropy \\
        \hline
        \myrowcolour \multirow{3}{*}{Kenya} & Average &  ${63.0 \pm 5.1}$ & \highest{71.6 \pm 3.7} & $ {73.9 \pm 4.4}$  \\
                            & Maximum  & $ 55.4 \pm 8.0$ & $ 58.9 \pm 16.7$ & $ 85.3 \pm 7.7$  \\
        \myrowcolour                    & Product & $ 50.0 \pm 0.0$ & $ 45.6 \pm 12.8$ & $ 99.7 \pm 9.9$ \\
                            & Concatenate & \highest{64.0 \pm 5.2} & $ 67.5 \pm 9.8$ & $ 79.8 \pm 0.4$ \\
                     \hline
        \myrowcolour \multirow{3}{*}{Togo} & Average & \highest{79.9 \pm 1.0} &  \highest{88.9 \pm 0.8} & $ {57.6\pm 4.1}$  \\
                            & Maximum & $ 64.7 \pm 9.0$ & $ 76.0 \pm 7.6$ & $ 86.9\pm 9.9$ \\
        \myrowcolour                    & Product  & $ 50.0 \pm 0.0$ & $ 49.9 \pm 8.0$ & $ 99.5\pm 0.5$  \\
                            & Concatenate & $ 79.5 \pm 1.9$ & $88.5 \pm 1.8$ & $ 59.2 \pm 3.7$ \\
                     \hline
        \myrowcolour \multirow{3}{*}{Global} & Average & $ 80.5\pm 1.8$ & $ 89.1 \pm 1.5$ & $ 54.0 \pm 3.3$  \\
                            & Maximum & $ 81.1\pm 1.1$ & $ 89.5\pm 1.0$ & $ {53.6\pm 2.7}$ \\
        \myrowcolour                    & Product  & $ 79.5\pm 1.2$ & $ 88.8\pm 0.9$ & $ 62.9 \pm 4.6$  \\
                            & Concatenate & \highest{81.3\pm 1.1} &  \highest{89.6\pm 1.1} & $ 53.8\pm 3.8$ \\
            \hline
    \end{tabular}
    \caption{Merge function comparison in the \textbf{Feature-S} method. The best average result for each test is underlined.}
    \label{tab:merge_func}
\end{table}
In Tab.~\ref{tab:gated_func} we compare different gate types in the feature-level fusion with gated modules (Feature-G). The standard gated module \cite{arevalo2020gatedmultimodalnetworksa} where the \textit{concatenation} of the view-representation is used to compute the weights is Gated-C, while Gated-A uses the \textit{average}. The GatedF uses a feature-specific weight in the gated module.
Similarly to the Feature-S method, the average obtains better performance than concatenation. Nevertheless, including the feature-specific weight improves overall.
Given that each feature might contain different information, the method allows more flexibility when combining the view's information.
\begin{table}[!t]
    \centering
    \small
    \begin{tabular}{cc|ccc} \hline
        Data & Type & AA & AUC & Entropy  \\
        \hline
        \myrowcolour \multirow{3}{*}{Kenya} & Gated-C  & $ 57.9\pm 6.1$ & $ 64.9 \pm 7.2$& $ 68.0\pm 7.0$  \\
                            & Gated-A & $ 58.5\pm 7.5$ & $ 66.4\pm 7.8$ & $ 72.3\pm 6.6$  \\
        \myrowcolour                    & GatedF-A & \highest{66.5 \pm 3.6} & \highest{71.8\pm 3.5} & $ {71.6\pm 5.8}$ \\
                     \hline
          & Gated-C  & $ 75.6\pm 5.8$ & $ 86.3\pm 2.4$ & $ 55.7\pm 13.6$  \\
        \myrowcolour   {Togo}                 & Gated-A & $ 77.1\pm 5.3$ & $ 87.3\pm 2.4$ & $ 55.0\pm 13.6$  \\
                            & GatedF-A &  \highest{78.1\pm 1.8} & \highest{87.6\pm 1.2} & $ {52.1\pm 7.8}$ \\
                     \hline
        \myrowcolour \multirow{3}{*}{Global} & Gated-C  & $ 80.3\pm 1.9$ & $ 88.8\pm 1.8$ & $ 53.3\pm 2.5$  \\
                            & Gated-A & $ 81.6\pm 0.7$ & $ 90.0\pm 0.6$ & $ 53.1\pm 3.5$  \\
        \myrowcolour                    & GatedF-A &  \highest{82.1\pm 1.4} & \highest{90.6\pm 1.1} & $ {51.1\pm 4.3}$ \\
            \hline
    \end{tabular}
    \caption{Gate types comparison in the \textbf{Feature-G} method. The best average result for each test is underlined.}
    \label{tab:gated_func}
\end{table}

\section{Final Remarks} \label{sec:conclu}
We present an extensive comparison of fusion methods in a crop classification benchmark, achieving state-of-the-art performance in the African testing regions without pre-training.
Even though our results are promising regarding different fusion strategies, there are some approaches that decrease the classification performance regarding individual views, while increasing prediction uncertainty.
The best performing fusion methods usually depend on the region where they are applied.
However, we suggest the following preliminary criterion.
When testing in a region with a large positive area (a higher percentage of positively labeled pixels in the region, e.g. Kenya/Global), the methods with feature-level exchange between views (Feature-G, Feature-S, Multi-Loss) are the best. 
Whereas, when testing in a region with a small positive area (e.g. Togo), the methods with feature-level exchange get worse, and exchanging at the output level becomes the best (Ensemble, Decision).
Anyhow, questions remain open whether the same results apply to other RS applications.

\bibliographystyle{IEEEbib}
\bibliography{refs}

\end{document}